\documentclass[journal]{journal}
\ifCLASSINFOpdf
\else
\fi
\usepackage{amsmath,graphicx,cite}
\usepackage{subfig}
\usepackage{amsmath}
\usepackage{amssymb}

\begin{document}
\title{Mushrooms Detection, Localization and 3D Pose Estimation using RGB-D Sensor for Robotic-picking Applications}

\author{Nathanael L. Baisa, Bashir Al-Diri

\thanks{Nathanael L. Baisa is with the School of Computer Science and Informatics, De Montfort University, Leicester LE1 9BH, UK. 
Email: nathanael.baisa@dmu.ac.uk.}
\thanks{Bashir Al-Diri is with the School of Computer Science, University of Lincoln, Lincoln LN6 7TS, UK. 
Email: baldiri@lincoln.ac.uk.}
%\thanks{Manuscript received April 19, 2005; revised January 11, 2007.}
}


\markboth{Journal of \LaTeX\ Class Files,~Vol.~6, No.~1, January~2007}%
{Shell \MakeLowercase{\textit{et al.}}: Bare Demo of IEEEtran.cls for Journals}



\maketitle
\thispagestyle{empty}

\begin{abstract}

In this paper, we propose mushrooms detection, localization and 3D pose estimation algorithm using RGB-D data acquired from a low-cost consumer RGB-D sensor. We use the RGB and depth information for different purposes. From RGB color, we first extract initial contour locations of the mushrooms and then provide both the initial contour locations and the original image to active contour for mushrooms segmentation. These segmented mushrooms are then used as input to a circular Hough transform for each mushroom detection including its center and radius. Once each mushroom's center position in the RGB image is known, we then use the depth information to locate it in 3D space i.e. in world coordinate system. In case of missing depth information at the detected center of each mushroom, we estimate from the nearest available depth information within the radius of each mushroom. We also estimate the 3D pose of each mushroom using a pre-prepared upright mushroom model. We use a global registration followed by local refine registration approach for this 3D pose estimation. From the estimated 3D pose, we use only the rotation part expressed in quaternion as an orientation of each mushroom. These estimated (X,Y,Z) positions, diameters and orientations of the mushrooms are used for robotic-picking applications. We carry out extensive experiments on both 3D printed and real mushrooms which show that our method has an interesting performance.

\end{abstract}

\begin{IEEEkeywords}
Mushrooms detection, Localization in 3D space, Active contour, Circular Hough transform, 3D pose estimation, RGB-D sensor
\end{IEEEkeywords}

% \fi
\IEEEpeerreviewmaketitle

\section{Introduction}

%The Mushroom Robo-Pict project will develop an automated robotic picking system for fresh mushroom crops. It addresses a key threat to the long-term future of the £120m UK mushroom industry arising from reduced labour availability. Picking labour current accounts for ca. 1/3 of production costs and this could all be removed through robotic harvesting. Previous work, across global centres, has hitherto failed to develop a fully robotic mushroom harvester.

Harvesting of fruits and vegetables such as grapes, mangoes, apples, kiwifruit, peaches, citrus, cherries, pears and mushrooms is highly labor intensive and poses human risk, for instance, ladder-related injuries that require a significant amount payment for compensation. It is also becoming less sustainable with increasing cost and decreasing availability of a skilled labor force. For instance, the \pounds 120m UK mushroom industry is facing challenges from reduced labour availability due to changes in demographics i.e. fewer younger people are entering the agricultural community whilst global food demand is ever expanding, and changes in national inward migration policies that may impact labour availability. Picking labour currently accounts for about 1/3 of production costs. To meet the increasing labor demand, to lower human risk of injuries, and to decrease the harvesting cost, automated harvesting system is crucial.

Many researchers and private companies have attempted to develop fully automated robotic system for harvesting of fruits and vegetables, however, no commercialized robotic harvester is available yet i.e. the need to develop robotic approaches to pick fresh fruits and vegetables is global. To develop a robotic harvester, the accuracy of detection, localization and 3D pose (rotation and translation) of fruits and vegetables is very important~\cite{GongAma15}\cite{PetKoi16}. However, there are many challenges to develop accurate detection, localization and 3D pose estimation algorithms such as changing illuminations, severe occlusions, various sizes and textures of fruits and vegetables, highly unstructured scenes, etc. Furthermore, these challenges cause robotic harvesting to be developed for specific data and designed for the task at hand.

There are some previous works of applying computer vision techniques for agriculture (agrovision) as reviewed in~\cite{KapBar12}\cite{GongAma15}. Some segmentation techniques for fruit detection are discussed in~\cite{SucJam16}, and some works are particularly tuned for a specific type of fruits such as grapes, mangoes, apples, kiwifruit, peaches, etc. Fruit detection, tracking, and 3D reconstruction for crop mapping and yield estimation was developed in~\cite{MooCha10}. These all previous works have focused mainly on 2D images. The recent introduction of cheap RGB-D sensors have opened a new door for computer vision researchers for many real applications. RGB-D sensors combine RGB color information with per-pixel depth information. Though sensors such as the Swiss Ranger SR4000 and Photonic Mixing Device (PMD) Tech products which can provide such data have existed for years, they are very costly. However, new consumer RGB-D sensors such as Kinect, Intel RealSense and Zed stereo camera are very cheap when compared to the aforementioned sensors. The applications and the challenges of Kinect sensor, particularly Kinect version 1 was explained in~\cite{CruzLuc12}. Both Kinect version 1 and 2 send out infrared laser light in order to measure the depth. The Kinect version 1 uses the structured-light-method to measure the depth with Pattern Projection principle. In this case, a known infrared (IR) pattern is projected into the scene and the depth is computed out of its distortion. However, the Kinect version 2 uses the Time-of-Flight (ToF) method and determines the depth by measuring the time the emitted light takes from ToF camera to the object and return. The Kinect version 2 provides more accurate depth information than the older version~\cite{WasStr17}\cite{YanZhaDon15}. For obtaining depth information from the Kinect version 2, measurable distance range is from 0.5m  to 4m~\cite{YanZhaDon15}. More importantly the depth accuracy of Kinect version 2 is almost constant over different distances~\cite{YanZhaDon15}. However, it is recommended to pre-heat the device for at least 25 min to obtain reliable results~\cite{WasStr17}. The ZED Stereo Camera, a depth sensor based on passive stereovision, observes objects from 0.5 to 20m. Intel RealSense cameras, both SR300 and D435, use infrared depth-sensing technology. D435 use active stereo infrared technology, which means the device has two infrared cameras as opposed to a single one on the SR300. SR300 is a short-range sensor (hence the model name) using coded-light method like Kinect version 1 whereas D435 is a longer-range sensor with a larger field of view as it uses depth stereo. The measurable distance range for SR300 is 0.2m - 1.5m whereas for D435 is 0.11m - 10m. The maximum range of D435 varies depending on calibration, scene, lighting condition, etc. The emergence of these low-cost RGB-D sensors motivated many computer vision researchers to use them for many applications such as object recognition~\cite{PraAld15}\cite{AndJos15}, object detection~\cite{XiaYun17}, pose estimation~\cite{SchBeh17}\cite{NguKoJeo15}\cite{ChoTagTuz12}, 3D reconstruction~\cite{IzadiKim11}, simultaneous localization and mapping (SLAM)~\cite{EndCre14}, etc.

However, to the best of our knowledge, only the work in~\cite{Mas13} gave attention to the application of computer vision for mushrooms harvesting. This work focused on a specific mushroom type, brown colour, and developed segmentation algorithm based on K-means. Though a good start, it has many limitations, for instance, the number of clusters needs to be predefined and the colour of the mushrooms to be applied to needs to be known in advance. It also did not include on any work on 3D pose estimation of the mushrooms. Our work takes one step forward by removing any prior knowledge about the mushrooms colour, illumination, etc as well as estimating depth information even if it is missing due to the inherent problems of the RGB-D sensors such as noise and holes. Basically the goal is to discriminate the mushrooms from the rest of the scene, to detect the presence of individual mushrooms, to localise them in space, and to estimate their 3D pose.

%In this work, we propose mushrooms detection, localization and 3D pose estimation algorithm for real-time robotic-picking applications. Accordingly, we make the following four contributions. First, we apply active contour for mushrooms segmentation under varying illumination conditions by providing an automatic initial contour locations. Second, we apply a circular Hough transform on the active contour-based segmented results for the detection of individual mushrooms of varying size. Third, we localize the mushrooms' centers by estimating from the nearest available depth information in case of missing. Fourth, we estimate the 3D pose of mushrooms using global registration followed by local refine registration approach. Finally, we make extensive experiments on both 3D printed and real mushrooms.

In this work, we propose mushrooms detection, localization and 3D pose estimation algorithm for real-time robotic-picking applications. Accordingly, our contributions can be summarized as follows. 

\begin{enumerate}
\item We segment mushrooms using active contour under varying illumination conditions by providing an automatic initial contour locations.
\item We detect individual mushrooms of varying size by providing the active contour-based segmented results to a circular Hough transform.
\item We localize the mushrooms' centers by estimating from the nearest available depth information in case of missing.
\item We estimate the 3D pose of mushrooms using global registration followed by local refine registration approach.
\item We make extensive experiments on both 3D printed and real mushrooms.
\end{enumerate}

The paper is organized as follows. The overview of our proposed method is given in section~\ref{sec:ProposedMethod}. In section~\ref{sec:Segmentation}, the segmentation method is described in detail. The mushrooms detection and localization approaches are explained in section~\ref{sec:MushroomsDetection} and section~\ref{sec:MushroomsLocalization}, respectively. In section~\ref{sec:MushroomsPose}, the 3D pose estimation is given. The experimental results are analyzed and compared in section~\ref{sec:ExperimentalResults}. The main conclusions and suggestions for future work are summarized in section~\ref{sec:Conclusion}.

\section{Overview of our Proposed Method} \label{sec:ProposedMethod}

\begin{figure*}[!htb]%[!htb] %\begin{figure*}
\begin{center}
   \includegraphics[width=1.0\linewidth]{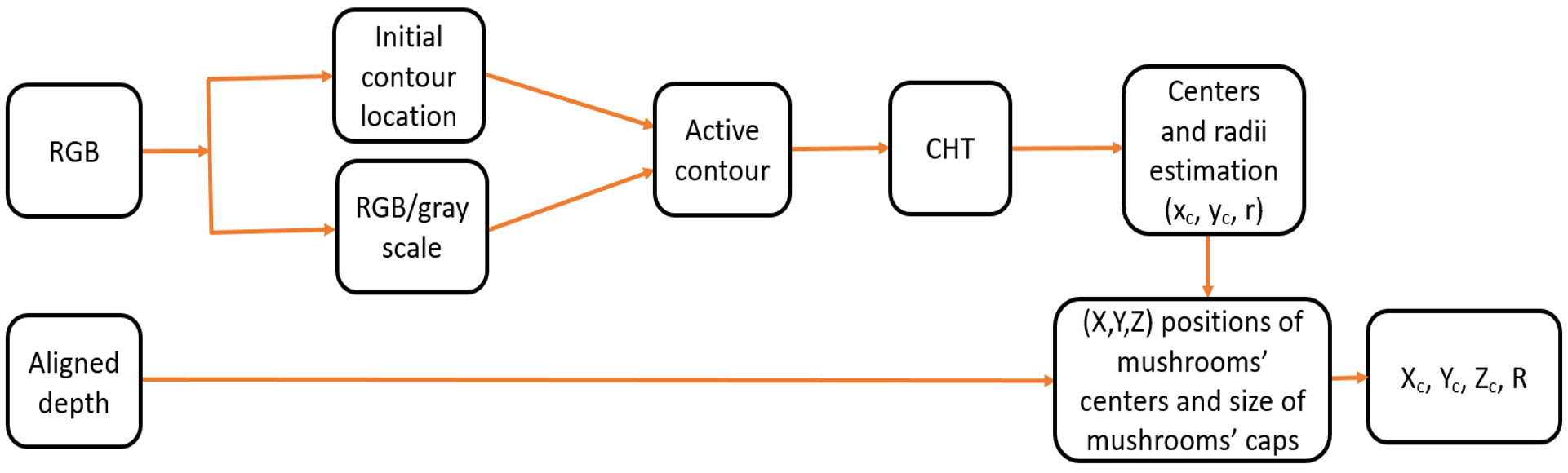}
\end{center}
   \caption{\small{Schematic diagram illustrating the proposed method for mushrooms segmentation, detection, and center (X,Y,Z) positions  and radii estimations using RGB-D data.}} %\vspace{-5mm}
\label{fig:ProposedMethod_DL}
\end{figure*}

\begin{figure*}[!htb]%[!htb] %\begin{figure*}
\begin{center}
   \includegraphics[width=1.0\linewidth]{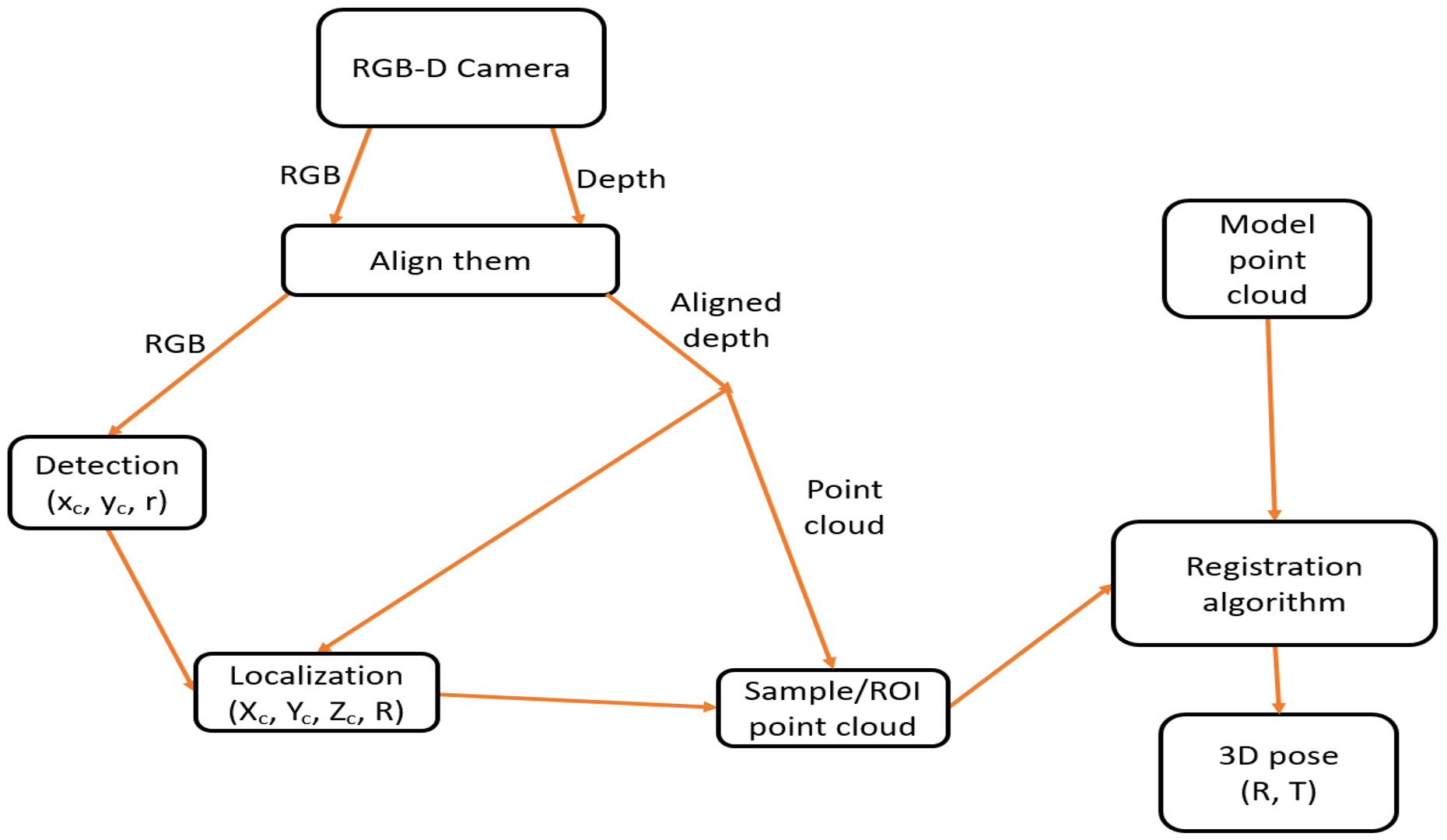}
\end{center}
   \caption{\small{Schematic diagram illustrating the proposed method containing all components: mushrooms detection, localization in 3D space, and 3D pose estimation using RGB-D data.}} %\vspace{-5mm}
\label{fig:ProposedMethod_DLP}
\end{figure*}

A schematic diagram of proposed detection and localization algorithm is given in Fig.~\ref{fig:ProposedMethod_DL}. It consists of three main components: segmentation using active contour, detection using a circular Hough transform and 3D localization using depth information. The overall diagram of the proposed algorithm including all components is shown in Fig.~\ref{fig:ProposedMethod_DLP}. Each of the components are discussed in detail as follows.

 % \end{figure*}
\noindent

 % \end{figure*}
\noindent

\section{Mushrooms Segmentation} \label{sec:Segmentation}

For mushrooms segmentation, we use an active contour method developed in~\cite{ChanVese01}, a type of active contour or snake usually known as Chan-Vase method. Basically active contour evolves the segmentation, splitting the image into foreground (object) and background regions, using an iterative process. This Chan-Vase method can segment objects whose boundaries are not necessarily defined by gradient or with very smooth boundaries. Starting from the initial contour location (from where the evolution starts) which can be anywhere in the image and not necessarily surround the objects of interest, this method can automatically segment the objects, in our case mushrooms. We use automatic thresholding to define the mask which is a binary image that specifies the initial state of the active contour. As shown in Fig.~\ref{fig:ProposedMethod_DL}, we use both RGB and graycale images independently as input to the active contour and we found that active contour works better when using grayscale image than the RGB image with a specified mask.

To state the mathematics behind this approach, let $\Omega$ be a bounded open subset of $\mathbb{R}^2$ with $\partial\Omega$ as boundary, $I_0: \overline{\Omega} \rightarrow \mathbb{R}$ be a given image, and $C$ be a parameterized curve. For the level set formulation of the used variational active contour model, the unknown variable $C$ can be replaced by the unknown variable $\phi$. The Heaviside function $H$ and the one-dimensional Dirac measure $\delta_0$ are used which can be defined, respectively, by

\begin{equation}
  H(x)=\begin{cases}
    1, & \text{if $z \geq 0$}\\
    0, & \text{if $z < 0$},
  \end{cases}
   \delta_0 = \frac{d}{dz} H(z)
\label{eqn:H}
\end{equation}
\noindent The energy functional $F(c_1, c_2, C) = F(c_1, c_2, \phi)$ is given by

\begin{equation}
\begin{array} {lll}  F(c_1, c_2, \phi) = & \mu (\int_{\Omega} \delta(\phi(x,y))|\nabla\phi(x,y)|dxdy)^p  \\& + \nu \int_{\Omega} H(\phi(x,y))dxdy \\& + \lambda_1\int_{\Omega} |I_0(x,y) - c_1|^2H(\phi(x,y))dxdy \\& + \lambda_2\int_{\Omega} |I_0(x,y) - c_2|^2(1 - H(\phi(x,y)))dxdy
\end{array}
\label{eqn:F}
\end{equation}
\noindent where the constants $c_1$, $c_2$, depending on $C$, are the averages of $I_0$ inside $C$ ($\phi \geq 0)$ and respectively outside $C$ ($\phi < 0$), and $\mu \geq 0$, $ \nu \geq 0$, $\lambda_1, \lambda_2 > 0, p \geq 1$ are fixed parameters. $\nabla\phi(x,y)$ is the gradient of $\phi(x,y)$. In our experiments, we set $\lambda_1, \lambda_2 = 1$, $\nu = 0$ and p = 1. The length parameter $\mu$ has a scaling role i.e. if we need to detect (segment) all or as many objects as possible and of any size, then it should be set to a small value and given as input together with the initial level set function $\phi_0$.

The solution of this model, $I$, is written using the level set formulation as

\begin{equation}
I (x,y) = c_1 H(\phi(x,y)) + c_2 (1 - H(\phi(x,y))), (x,y) \in \overline{\Omega}
\label{eq:u}
\end{equation}
\noindent In the equation~(\ref{eqn:F}), the first term can be thought as a penalty on the total length of the edge contour for a given segmentation, the second term is a penalty on the total area of the foreground region found by the segmentation, the third term is proportional to the variance of the image gray level in the foreground region, and the fourth term does the same for background region. The level set function $\phi$ is used only to represent the curve and the energy $F$ is minimized to find its solution. A global minimizer can be obtained, independently of the position of the initial curve, which allows to automatically detect interior contours. This minimization problem could also be solved numerically, using a similar finite differences scheme. The good advantage of this model is that it is not dependent on an edge-function to stop the evolving curve on the desired boundary i.e. we can detect objects whose boundaries are not necessarily defined by gradient or with very smooth boundaries. It is also possible automatically detect interior contours starting with only one initial curve. The position of the initial curve can be anywhere in the image, and it does not necessarily surround the objects to be detected.

A version of the Chan-Vese algorithm that uses morphological operators instead of solving a partial differential equation (PDE) for the evolution of the contour can also be used~\cite{MarBauAlv14}. The set of morphological operators used in this algorithm are proved to be infinitesimally equivalent to the Chan-Vese PDE. %However, they overcome the numerical stability issues typically found in PDEs, and are computationally faster.

\section{Mushrooms Detection} \label{sec:MushroomsDetection}

There are many features such as color, geometric, texture or their combination that can be used in machine vision for fruits and vegetables detection. Geometric measures such as shape and size provide a good set of distinct features of fruits and vegetables. In our case, the shape of mushrooms, particularly mushrooms caps which are circles, gives a good discriminative feature to use for their detection. Accordingly, we apply a Circular Hough Transform (CHT)~\cite{AthKer99}, which can detect circles of varying size, to the segmented (binary) image for mushrooms caps detection. We first apply opening morphological operation with elliptical structuring element (kernel) of (10,10) on the segmented images to remove small objects (noise) and then apply CHT for detection of the mushrooms. Given a radius range [Rmin, Rmax], the CHT algorithm detects the mushrooms centers and radii of varying size in the image.

For the CHT computation, a separate circle filter is used for detecting each radius of circle which basically forms the 3-dimensional parameter space. The two dimensions represent the position of the circle center ($c_x,c_y$) and the third is radius $R$.

\begin{equation}
(x-c_x)^2 + (y-c_y)^2 = R^2
\label{eq:CHT}
\end{equation}
\noindent The CHT developed in~\cite{AthKer99} which has shown a specific combination of modifications to the CHT is equivalent to applying a scale invariant kernel
operator. Accordingly, the invariance kernel for scaling in two dimensions is
derived from the scaling group

\begin{equation}
\begin{array} {lll}
x' &= \pi x \\
y' &= \pi y
\end{array}
\label{eq:xy}
\end{equation}
\noindent where (x, y) is the original coordinate, (x', y') the transformed coordinate, and $\pi$ the scaling factor. The Lie derivative,
or generator for this group is given as
\begin{equation}
L_s = x \frac{\partial}{\partial x} + y \frac{\partial}{\partial y}
\label{eq:Ls}
\end{equation}
\noindent Then the invariance kernel for scaling and rotation in 2D is

\begin{equation}
K_{rs}(k,l) = \frac{1}{r^2} \exp[i(k\theta + l\ln r)]
\label{eq:Krs}
\end{equation}
\noindent A circularly symmetric member of this family is chosen by setting $k=0$ i.e. $K_{rs}(0,l)$ and then the first radial component is considered. A change in circle size is transformed by this into a shift in log size, i.e. into a translation $r' = \pi r \rightarrow \log r' = \log r + \log \pi$. The kernel is shift variant so that it can be applied to every point in the image in the form of a convolution. The resulting image has the convolution result $I(x,y) \circledast K_{rs}(0,l)$ at each point. This gives a response that is a peak at the centre of the circle, the magnitude of this peak is invariant to the size of the circle, and phase at the peak gives a measure of the circle size. This approach has a reasonable performance and is used in our approach as it allows to detect circles of varying size.

\section{Localization of Mushrooms' Centers} \label{sec:MushroomsLocalization}

\begin{figure}
\centering
\begin{minipage}[c]{0.48\textwidth} %0.48
\centering
    \includegraphics[width=0.70\linewidth]{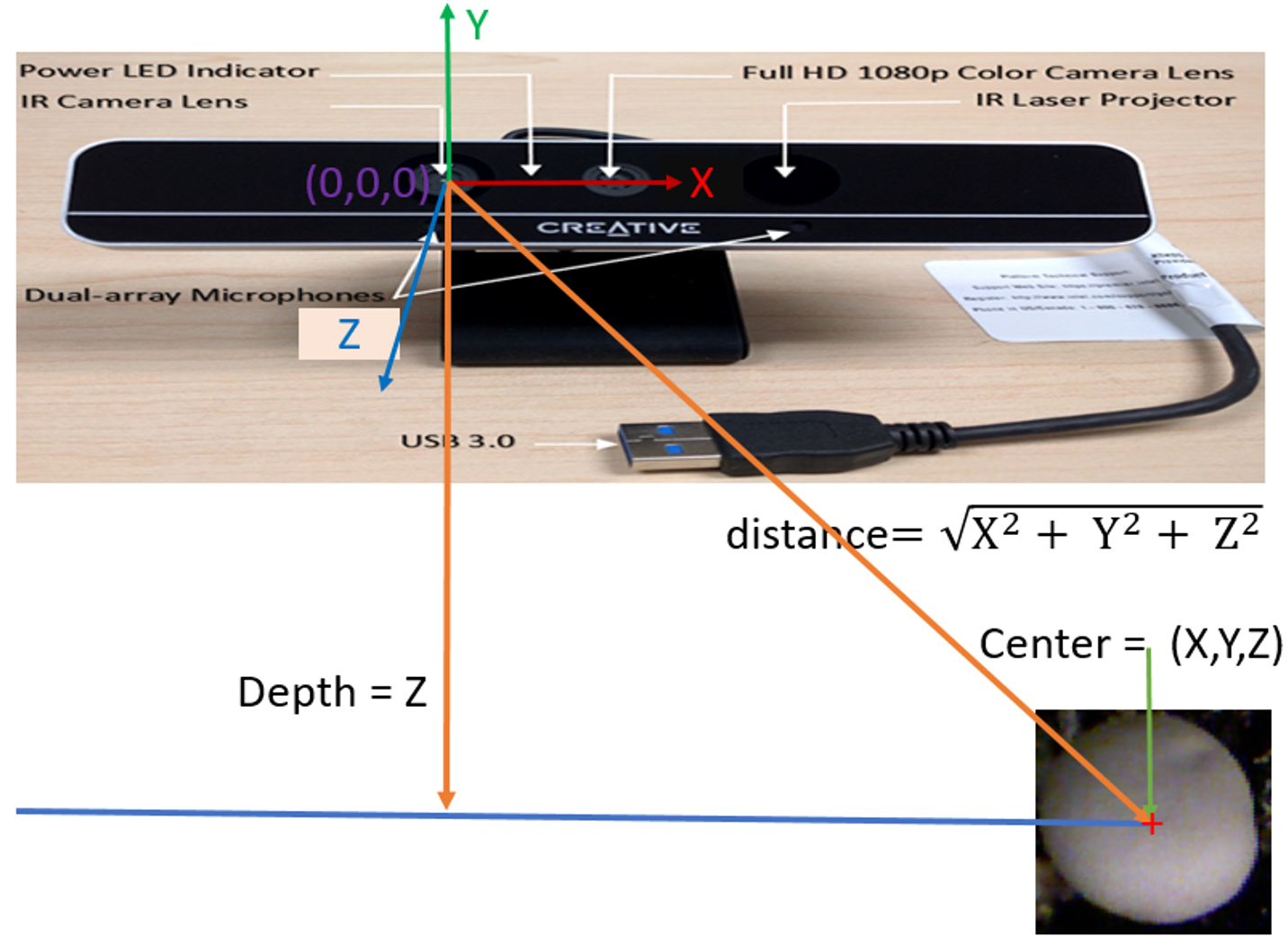}  %0.7
    \caption{\small{Intel RealSense SR300 configuration showing depth and distance to the center position (X,Y,Z) of a mushroom cap. The origin (X=0, Y=0, Z=0) is located at the center of the depth sensor.}}
    \label{fig:SR300Config}
\end{minipage}
\end{figure}

Once mushrooms caps are detected, it is necessary to locate them in 3D space i.e. in the world coordinate system (X,Y,Z), and then these 3D coordinates of mushrooms centers with other parameters can be given as inputs to the robot-end effector for robotic-picking applications. To locate the mushrooms centers in 3D space, we use the depth information from the Intel RealSense SR300 due to its short-range sensing capability which fits well for our application (mushrooms needs to be sensed in less than 20cm distance between mushroom beds). In our experiment, SR300 gives much higher point cloud quality than the D435. We use the color and depth resolutions of $640\times480$. As shown in Fig.~\ref{fig:SR300Config}, the depth (Z) is the perpendicular distance from the sensor plane to the object of interest plane. The actual distance (d) of the object from the sensor is given by taking into account all X, Y and Z coordinates as

\begin{equation}
d = \sqrt{X^{2} + Y^{2} + Z^{2}}
\label{eq:distance}
\end{equation}
\noindent where d is the distance between the center of the depth sensor and the center coordinate of each individual mushroom cap (object of interest) with the fact that the origin (X=0, Y=0, Z=0) is located at the center of the depth sensor.

However, missing depth information (holes) and noise are some of the inherent problems in the RGB-D sensors such as the Intel RealSense sensor in general though some postprocessing alleviates them. In case of missing depth information at the detected center of each mushroom, we estimate from the nearest available depth information within the radius of each mushroom. To make robust to noise, we estimate (X,Y,Z) positions by taking average of the nearest available coordinates at which depth information is available.

\noindent

\section{Mushrooms 3D Pose Estimation} \label{sec:MushroomsPose}

The orientation estimation of each mushroom is also important for successful picking without harming it. Therefore, we propose a 3D pose (rotation and translation) estimation algorithm based on matching or registration of point clouds of a model with a sample (an interested mushroom). We prepare one model particularly upright mushroom model. We propose a method based on global registration followed by local registration. We use Open3D for our implementation~\cite{ZhoParKol18}.

\subsection{Global registration}  This registration method is used to estimate a rough alignment of two point clouds (model and sample) using sparse correspondences. Basically the dense point clouds are first downsampled and then 3D geometric descriptors are extracted from these downsampled (subsampled) point clouds. We use Fast Point Feature Histograms (FPFH) as the 3D geometric descriptors~\cite{RusBloBee09}. The FPFH feature is a 33-dimensional vector that describes the local geometric property of a point. Sparse correspondences are computed using the 3D geometric descriptors and a rough alignment of these point clouds are estimated using either sampling-based global registration methods such as RANSAC~\cite{HolIchTom15} or a single objective function optimization based global registration methods~\cite{ZhoParKol16}; the latter is used in our case as it is faster.

\subsection{Local registration}  Local registration is used as a local refinement by beginning with a rough initial alignment (obtained from the global registration stage). This produces a tight registration based on the original undownsampled point clouds i.e. while global registration methods operate on sparse candidate correspondences using 3D geometric descriptors, local registration methods operate on dense correspondences. We use Iterative Closest Point (ICP) algorithm~\cite{pomColSie13} for the local refinement stage. Even though the two variants~\cite{RusLev01}, point-to-point and point-to-plane based ICP local registration, work well in our experiment, we use the point-to-point method for our evaluation. Generally, the ICP algorithm iterates over two steps
\begin{enumerate}
  \item Find correspondence set $K=\{(p,q)\}$ from target point cloud P, and source point cloud $Q$ transformed with current transformation matrix $T$.
  \item Update the transformation $T$ by minimizing an objective function $E(T)$ defined over the correspondence set $K$.
\end{enumerate}
Basically, different variants of ICP use different objective functions $E(T)$~\cite{BesMcK92}\cite{CheMed92}\cite{ParZhoKol17}. We use the point-to-point ICP algorithm in~\cite{BesMcK92} using objective function

\begin{equation}
 E(T) = \sum_{(p,q)\in K} {||p-Tq||^2}
\label{eq:ObjectiveFunction}
\end{equation}
This aligns the source point clouds to the target point clouds. We use our upright mushroom model as the source and the mushroom sample as the target point cloud. This allows as to get the transformation from our upright mushroom model to each mushroom sample so that we can know how much the robot needs to transform itself for picking each mushroom.

The outputs of the last local registration stage are rotation and translation. Since we locate each mushroom using depth information, as discussed in section~\ref{sec:MushroomsLocalization}, only the rotation part of the obtained 3D pose is used. This rotation matrix (R) is converted to quaternion ($[q_x, q_y, q_z, q_w]$) and then is used as the orientation of each mushroom which can be used by the robot with the other parameters to effectively pick it. Basically, this obtained rotation matrix contains all rotation components i.e. rotation about X-axis (Roll), rotation about Y-axis (Pitch) and rotation about Z-axis (Yaw). The rotation component about the Z-axis is not useful as the robot can freely move in this axis for successful picking of each mushroom. For the evaluation purpose, the rotation component of the Z-axis can normally be cancelled by multiplying the obtained rotation matrix from the 3D pose estimation method with the unit normal vector along the Z-axis i.e. $\vec{n}_z = [0, 0, 1]$. However, this is only possible, in our case, if the mushroom model we use perfectly aligns with the Z-axis which is not always the case. Therefore, we use the unit normal ground truth vector of the mushroom model which we computed by ourselves to effectively cancel out the rotation around the Z-axis. This gives us the normal orientation vector of each mushroom which can be compared with the ground truth normal orientation vector of each mushroom to know the performance of our approach; please refer to section~\ref{sec:ExperimentalResults} for the quantitative and qualitative results.

%The rotation component about the Z-axis is not useful as the robot needs only orientation of each mushroom for successful picking, in our application. Thus, the rotation component of the Z-axis can normally be cancelled by multiplying the obtained rotation matrix from the 3D pose estimation method with the unit normal vector along the Z-axis i.e. $\vec{n}_z = [0, 0, 1]$. However, this is only possible, in our case, if the mushroom model we use perfectly aligns with the Z-axis which is not always the case. Therefore, we use the unit normal ground truth vector of the mushroom model which we computed by ourselves to effectively cancel out the rotation around the Z-axis. This gives us the orientation of each mushroom which can be used by the robot with the other parameters to effectively pick it.

\section{Experimental Results} \label{sec:ExperimentalResults}

\begin{figure*} [htb] %[t]%[!h]
  %\centering
  \begin{center}
   \subfloat[\scriptsize{Original image}]
  {\label{fig:3DprintedOriginal} \includegraphics[width=0.50\textwidth]{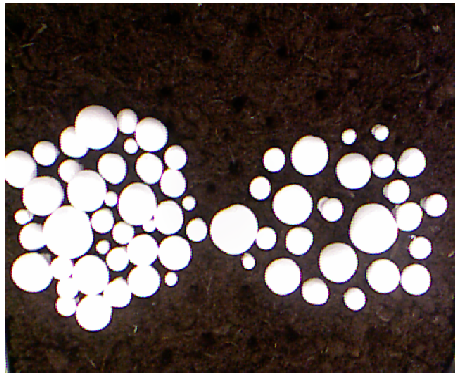}} %\\ %height=0.31
  \subfloat[\scriptsize{Initial contour location}]
  {\label{fig:3DprintedInitialC} \includegraphics[width=0.50\textwidth]{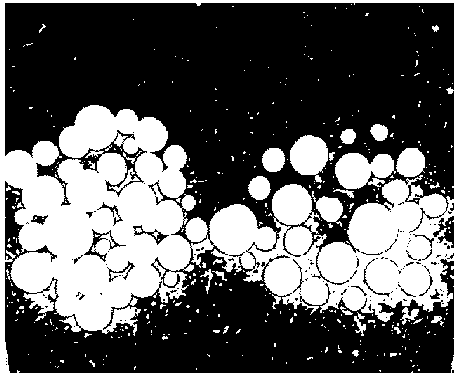}} \\%\\
  \subfloat[\scriptsize{Active contour and then morphological opening}]
  {\label{fig:3DprintedSegment} \includegraphics[width=0.50\textwidth]{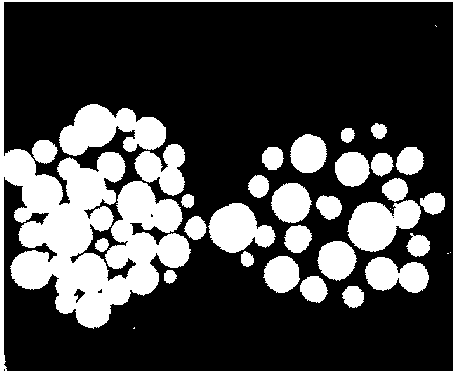}}  %\\
    \subfloat[\scriptsize{Circular Hough Transform (CHT)}]
  {\label{fig:3DprintedCHT} \includegraphics[width=0.50\textwidth]{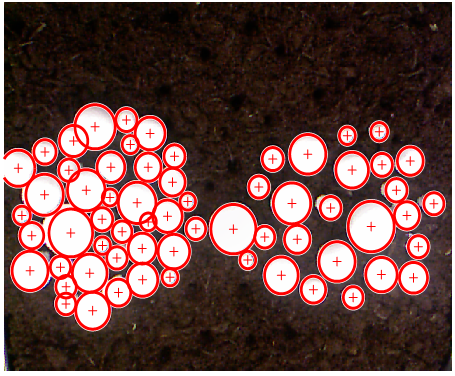}}
  \end{center}
   \caption{\small{Experimental results using 3D printed mushrooms. The '+' sign in (d) shows the center of each mushroom.}}
  \label{fig:3Dprinted}
\end{figure*}

\begin{figure*} [htb] %[t]%[!h]
  %\centering
  \begin{center}
     \subfloat[\scriptsize{Original image}]
  {\label{fig:RealMushroom8original} \includegraphics[width=0.50\textwidth]{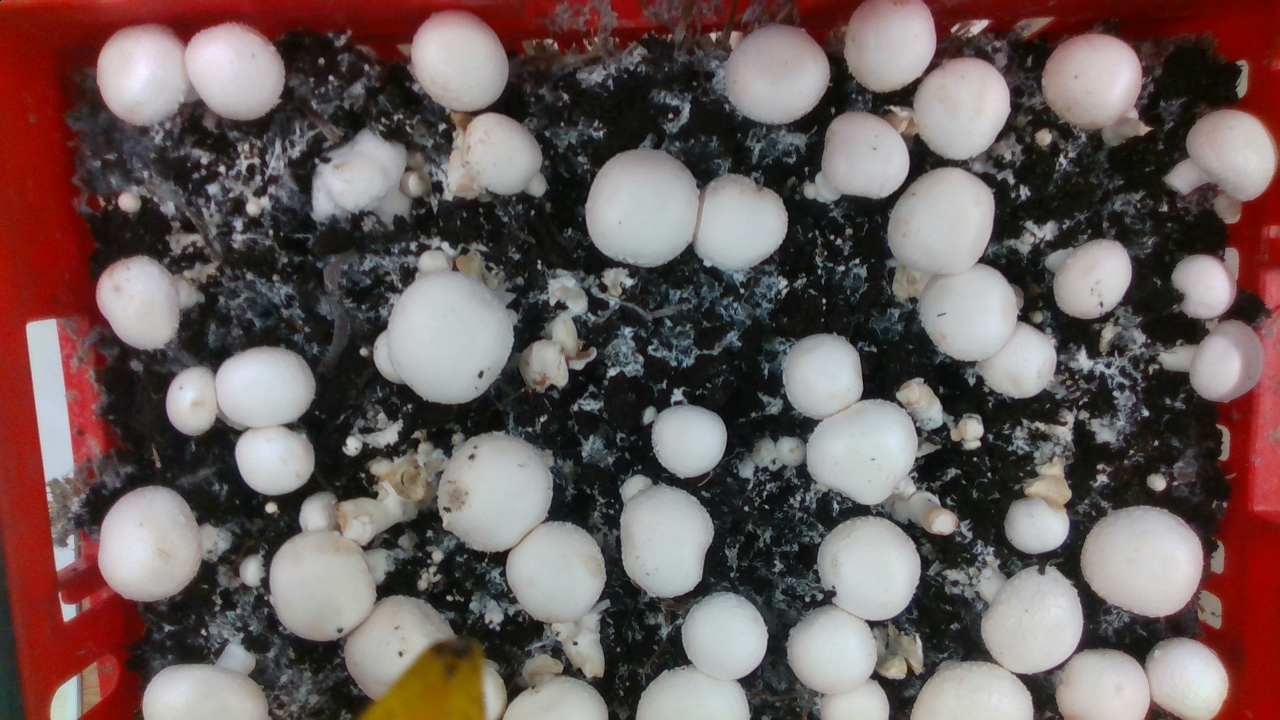}} %\\ %height=0.272
     \subfloat[\scriptsize{Segmentation using active contour}]
  {\label{fig:RealMushroom8AC} \includegraphics[width=0.50\textwidth]{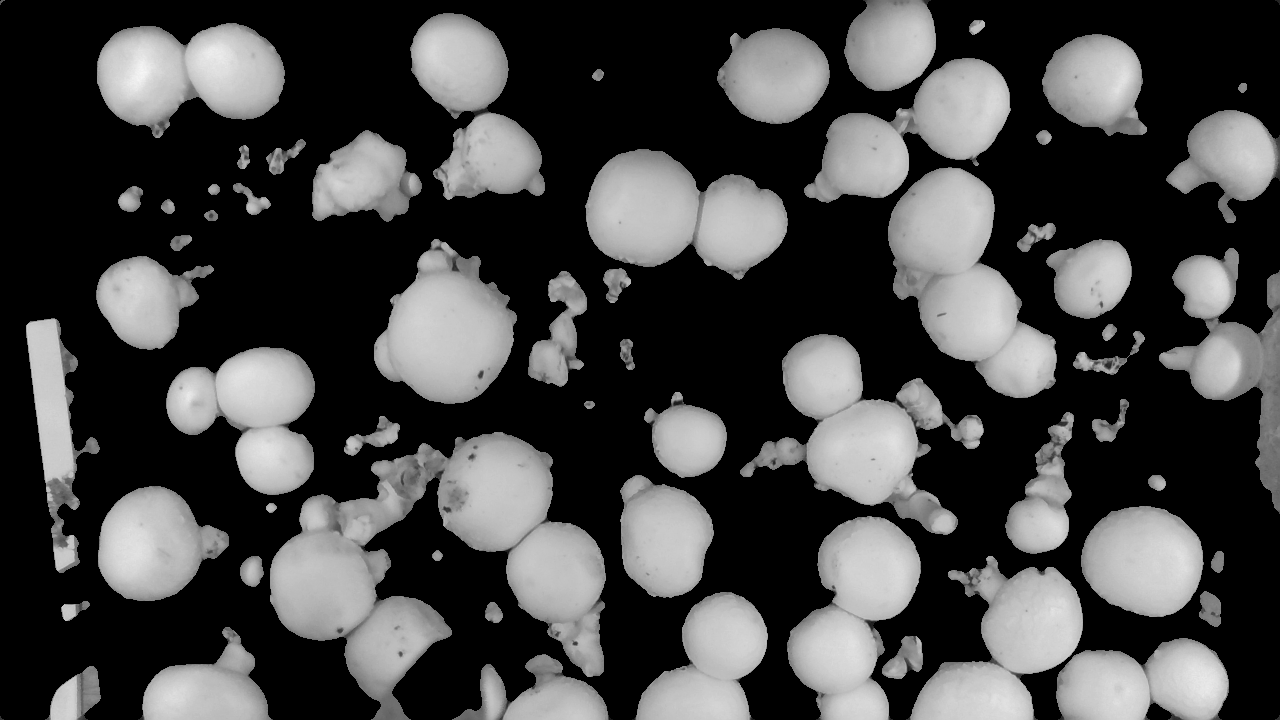}} \\ %height=0.272
    \subfloat[\scriptsize{Active contour and then morphological opening}]
  {\label{fig:RealMushroom8ACopening} \includegraphics[width=0.50\textwidth]{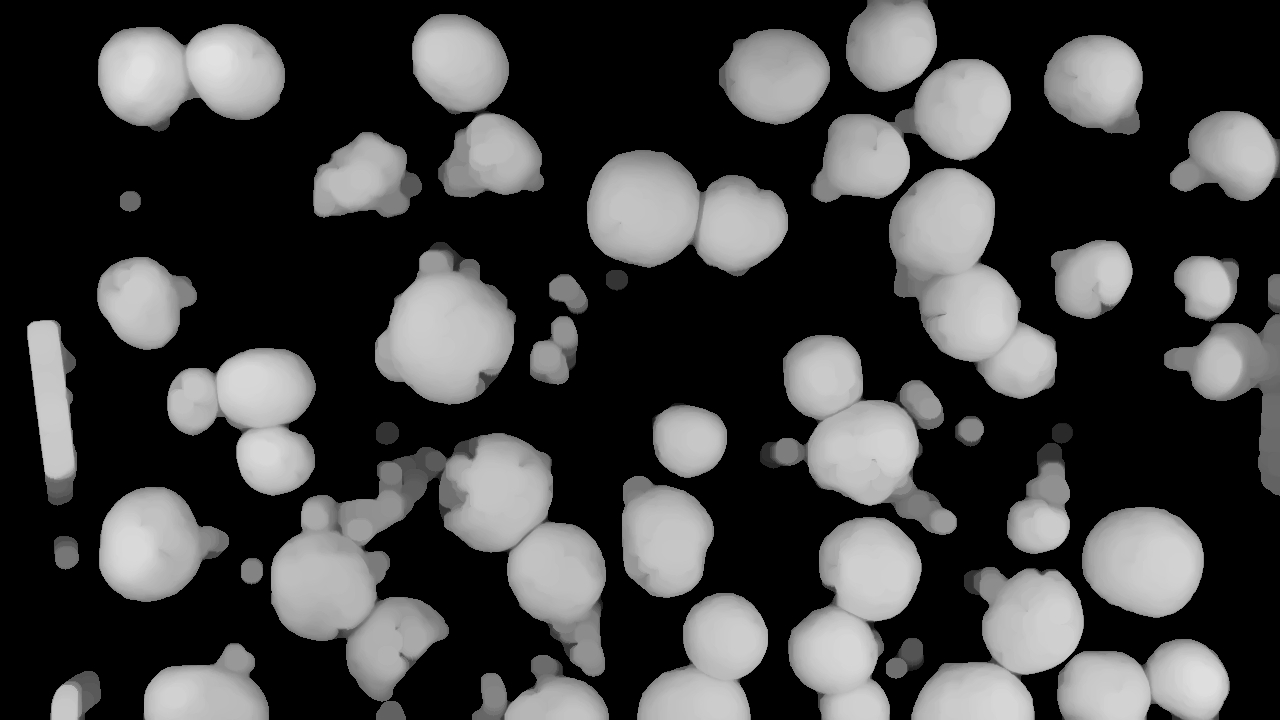}} %\\ %height=0.272
    \subfloat[\scriptsize{Detected mushrooms using CHT}]
  {\label{fig:RealMushroom8CHT} \includegraphics[width=0.50\textwidth]{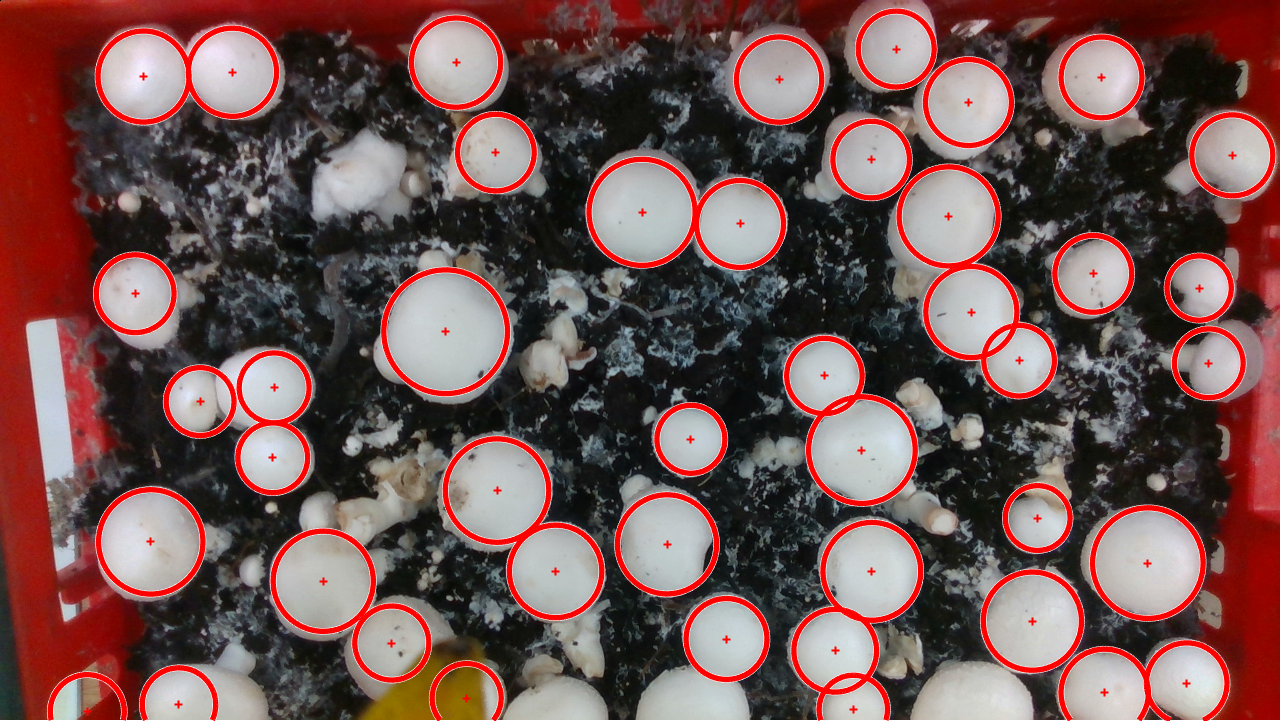}} %\\ %height=0.272
  \end{center}
   \caption{\small{Experimental results using real mushrooms. The '+' sign in (d) shows the center of each mushroom.}}
  \label{fig:RealMushroomSample2}
\end{figure*}

\begin{figure*} [htb] %[t]%[!h]
  %\centering
  \begin{center}
   \subfloat[\scriptsize{Original image}]
  {\label{fig:Real2Orignial} \includegraphics[width=0.50\textwidth]{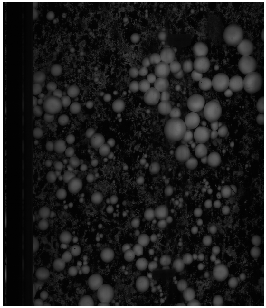}} %\\ %height=0.31
 % \subfloat[\scriptsize{Initial contour location}]
%  {\label{fig:Real2InitialC} \includegraphics[height=0.31\textwidth]{bashir2InitialC.png}} \\%\\
%  \subfloat[\scriptsize{Segmented image using active contour}]
%  {\label{fig:Real2Segment} \includegraphics[height=0.31\textwidth]{bashir2Segmented.png}}  %\\
    \subfloat[\scriptsize{Detected mushrooms using CHT}]
  {\label{fig:Real2CHT} \includegraphics[width=0.50\textwidth]{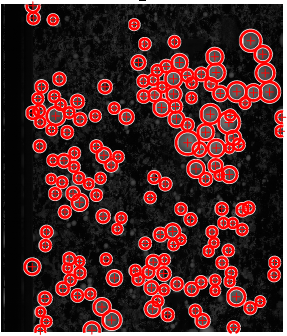}}
  \end{center}
   \caption{\small{Experimental results using real mushrooms. The '+' sign in (b) shows the center of each mushroom.}}
  \label{fig:RealBashir2}
\end{figure*}

\begin{figure*} [htb] %[t]%[!h]
  %\centering
  \begin{center}
   \subfloat[\scriptsize{Original image}]
  {\label{fig:Real3Orignial} \includegraphics[width=0.50\textwidth]{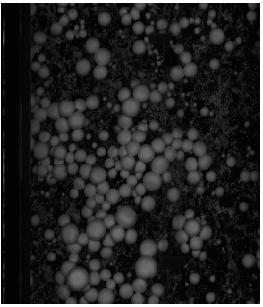}} %\\ %height=0.31
  %\subfloat[\scriptsize{Initial contour location}]
%  {\label{fig:Real3InitialC} \includegraphics[height=0.31\textwidth]{bashir3InitialC.png}} \\%\\
%  \subfloat[\scriptsize{Segmented image using active contour}]
%  {\label{fig:Real3Segment} \includegraphics[height=0.31\textwidth]{bashir3Segmented.png}}  %\\
    \subfloat[\scriptsize{Detected mushrooms using CHT}]
  {\label{fig:Real3CHT} \includegraphics[width=0.50\textwidth]{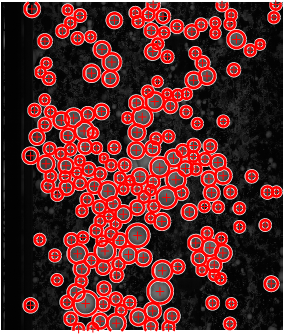}}
  \end{center}
   \caption{\small{Experimental results using real mushrooms. The '+' sign in (b) shows the center of each mushroom.}}
  \label{fig:RealBashir3}
\end{figure*}

We qualitatively and quantitatively evaluate our method on both 3D printed and real mushrooms. Since mushrooms grow fast and destroy after a few days, we use the 3D printed version to do experiment in our Lab. As shown in Fig.~\ref{fig:3Dprinted}, given the initial contour locations of the 3D printed mushrooms, active contour gives a good segmentation result and CHT also detect all of the mushrooms. This algorithm also shows similar performance on real mushrooms obtained from a mushroom farm as shown in Fig.~\ref{fig:RealMushroomSample2}, Fig.~\ref{fig:RealBashir2} and Fig.~\ref{fig:RealBashir3}. The method detects almost all mushrooms of varying size.

We use recall and precision for quantitative evaluations of the mushroom detection algorithm on the real mushrooms. Recall (also called sensitivity,  true positive rate, or probability of detection) measures the proportion of the mushrooms that are correctly detected. On the other hand, precision (also called the positive predictive value) measures the proportion of actual positives (correctly detected mushrooms) in the data. They are formulated as
\begin{equation}
Recall = \frac{TP}{TP + FN}
\label{eq:Sensitivity}
\end{equation}
\noindent where $TP$ is the number of true positives, and $FN$ is the number of false negatives (miss-detections).

\begin{equation}
Precision = \frac{TP}{TP + FP}
\label{eq:Sensitivity}
\end{equation}
\noindent where $FP$ is the number of false positives (false alarms). Any detections which have intersection-over-union of greater than or equal to 0.5 with the ground truth are treated as true positives whereas those with less than 0.5 are treated as false positives. Any ground truths which have intersection-over-union of less than 0.5 with detection are treated as false negatives (miss-detections). We use many captured real mushrooms from the mushroom farm (20 randomly selected from different images are used for our evaluations). The evaluations of our proposed detection algorithm on these real mushrooms provide sensitivity of 99.29\% and precision of 98.99\%.  The $F_1$-score (F-score or F-measure) can be used as a single measure of performance. The F-score is the harmonic mean of precision and recall:
\begin{equation}
\text{F-score} = 2 \times \frac{Recall \times Precision}{Recall + Precision}
\label{eq:Fscore}
\end{equation}
\noindent The evaluations of our proposed detection algorithm gives F-score of 99.14\%.

For this experiment, the range of radii used for CHT algorithm is [8 38] pixels (range of diameters is [16 76] pixels) i.e. mushrooms smaller than radius of 8 pixels or greater than radius of 38 pixels cannot be detected and are not included into our evaluations. This radius range was set empirically for our dataset; no mushroom with radius greater than 38 pixels was observed in the given images i.e. we set the maximum radius so that the algorithm can detect the biggest mushroom available. The mushrooms smaller than 8 pixels are uninterested ones (in our case) and are treated as true negatives. % TP = 2942, TN = 584, FP = 30, FN = 21

\noindent

%\begin{figure*} [htb] %[t]%[!h]
%  %\centering
%  \begin{center}
%   %\subfloat[\scriptsize{Original image}]
%  {\label{fig:RealMushroom} \includegraphics[width=1.0\textwidth]{real_mushrooms_sample2.png}} %\\ %height=0.272
%  \end{center}
%   \caption{\small{Experimental results using real mushrooms. }}
%  \label{fig:RealMushroomSample}
%\end{figure*}
%\noindent

\noindent

\noindent

\noindent

We also evaluate quantitatively the performance of the depth measurement of the Intel RealSense SR300. We determine the ground truth depth values by physically measuring the perpendicular distance from the perpendicularly downward facing Intel RealSense SR300 to the planar surface. The depth accuracy is the difference between the true depth value and the average value of the measured depth values corresponding to a planar surface located in front of the SR300. The depth accuracy can be computed using the following equation

\begin{equation}
Depth~Accuracy = GT_d - mean(M_d)
\label{eq:depthAccuracy}
\end{equation}
\noindent where $GT_d$ is the ground truth depth value between the camera plane and the surface plane whereas $mean(M_d)$ is the mean of the measured depth values over central pixels. In this case, we constrain the evaluation to the central pixels of $31 \times 31$ window around the camera center for reliability. The method in~\cite{YanZhaDon15} also uses central pixels for depth accuracy evaluation but for Kinect version 2. As given in Table~\ref{tbl:DepthAccuracy}, we also evaluate the standard deviation and range of the measured depth values over these central pixels. The accuracy of the SR300 is almost constant (around -3mm in our case) with a very good accuracy.

\begin{table*}[!htb]%[!h]%[tb]
\begin{center}
\begin{tabular}{|l|c|c|c|r|}
\hline
Ground Truth & Mean Depth & Std & Range & Offset (accuracy)\\
\hline\hline
79.30cm & 79.58cm & 1.1mm & 5mm & -2.8mm \\
71.78cm & 72.08cm & 1.1mm & 5mm & -3.0mm\\
58.55cm & 58.84cm & 1.0mm & 5mm & -2.9mm \\
50.83cm & 51.10cm & 1.1mm & 5mm & -2.7mm \\
\hline
\end{tabular}
\end{center}
\caption{\small{Evaluation of the distance between the Intel RealSense camera plane and the planar surface. Ground truth of depth values, mean of the measured depth values, standard deviation of the measured depth values, range of the measured depth values, and offset or error of the measured depth values from SR300.}}
\label{tbl:DepthAccuracy}
\end{table*}
\noindent

\begin{figure*} [htb] %[t]%[!h]
  %\centering
  \begin{center}
   %\subfloat[\scriptsize{Original image}]
%  {\label{fig:3Dprinted3Doriginal} \includegraphics[height=0.38\textwidth]{image3DprintedOriginal.png}} \\ %height=0.272
  \subfloat[\scriptsize{Depth (Z) retrieval from RGB-D data}]
  {\label{fig:3Dprinted3Dread} \includegraphics[height=0.38\textwidth]{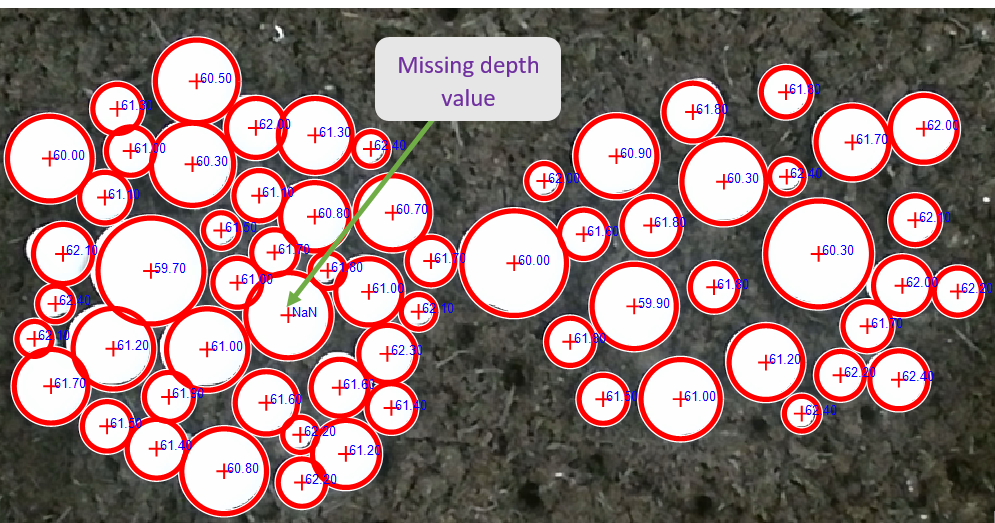}} \\
  \subfloat[\scriptsize{Depth (Z) and distance (d) estimation from nearest neighbours}]
  {\label{fig:3Dprinted3Destimate} \includegraphics[height=0.38\textwidth]{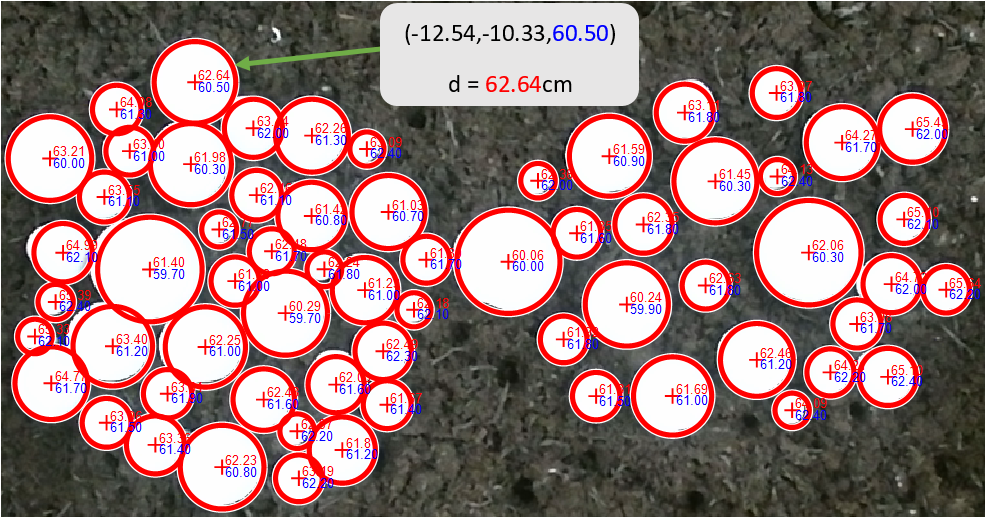}}  %\\  %0.2
  \end{center}
   \caption{\small{Mushrooms' center localization. In (b), the numbers in blue show the depth (Z) and the numbers in red show the distance from the sensor to each mushroom center. We ignore the X and Y components to display for clarity except for one sample for which we show both (X,Y,Z) position and distance (d) in centimeter.  The '+' sign in (a) and (b) shows the center of each mushroom. The missing depth information in (a) is estimated from the average of the nearest available depth information and is shown in (b).}}
  \label{fig:3Dprinted3Dlocalization}
\end{figure*}

\begin{figure*} [htb] %[t]%[!h]
  %\centering
  \begin{center}
   %\subfloat[\scriptsize{Original image}]
  {\label{fig:3Dprinted3DestimateIntelRealSense} \includegraphics[height=0.38\textwidth]{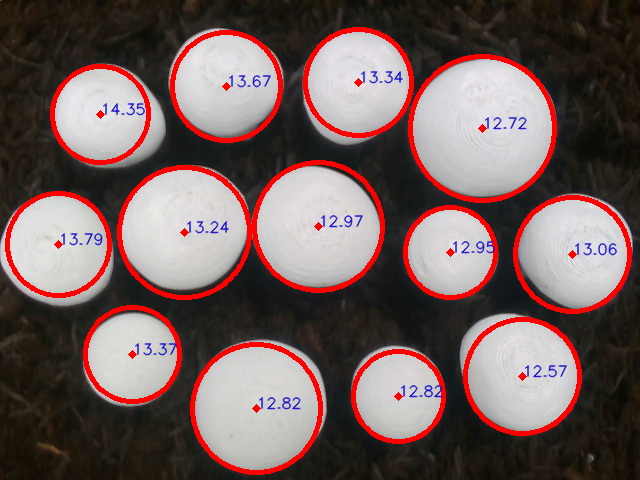}}  %height=0.272
  \end{center}
   \caption{\small{Mushrooms' center localization. The numbers in blue show the depth (Z). Sample result for the 3D printed mushrooms with the measured depth (Z) values in cm which are in 10's (This is not possible, for instance, using Kinect 2). The dot (.) sign shows the center of each mushroom.}}
  \label{fig:3Dprinted3DlocalizationIntelRealSense}
\end{figure*}

The localization of each mushroom center in 3D space is also shown in Fig.~\ref{fig:3Dprinted3Dlocalization}. The direct depth reading from the Intel RealSense is shown in Fig.~\ref{fig:3Dprinted3Dread}. As can be observed from this figure, 'NaN' is displayed for showing missing of depth information. Though depth information is obtained for many mushrooms centers in this experiment, depth information can sometimes be missing for some mushrooms centers. In this experiment, depth information is missing for one mushroom center as shown in Fig.~\ref{fig:3Dprinted3Dread}. We then estimate the 3D location of the mushroom center from the nearest available depth information. We take the average of the nearest available (X,Y,Z) positions which are obtained from nearest available depth information with the constraint that this nearest available depth information must be located within the radius of each mushroom. In this experiment, the nearest depth information is obtained within a few pixels distance from the center of the mushroom. %50\% of each radius.
We take the average of the nearest available depth information for better accuracy i.e. robust to noise. Accordingly, we only display the depth (Z) in blue and the distance (d) in red; we did not display the other coordinates (X and Y) for clarity. We show both (X,Y,Z) position and distance (d) from the depth sensor to the center of the mushroom for one sample as a typical example in Fig.~\ref{fig:3Dprinted3Destimate}. The depth accuracy of the mushrooms caps is in agreement with the depth accuracy given in Table~\ref{tbl:DepthAccuracy}. We measure the ground truth depth values of about 20 mushrooms caps and compare with their corresponding measured depth values; the mean accuracy (offset) obtained is about -2.77mm. The Intel RealSense camera can sense nearer objects than the other RGB-D sensors as shown in Fig.~\ref{fig:3Dprinted3DlocalizationIntelRealSense}; this is the reason we choose this sensor for our application. %, standard deviation of 1.3mm and range of 5mm.
% md = [59.9 61.5 61.8 61.6 62.0 60.0 61.70 60.7 62.4 61.0 62.1 62.3 61.4]
% gt = [59.6 61.1 61.2 61.4 61.7 59.7 61.6 60.5 62.3 60.6 61.9 62.0 61.2]
As can be seen from Fig.~\ref{fig:3Dprinted} and Fig.~\ref{fig:3Dprinted3Dlocalization}, the detection algorithm performs well on 3D printed mushrooms with recall and precision of unity.

%\begin{figure*} [htb] %[t]%[!h]
%  %\centering
%  \begin{center}
%   \subfloat[\scriptsize{Original image}]
%  {\label{fig:3Dprinted3Doriginal} \includegraphics[height=0.38\textwidth]{imageV2Original2.png}} \\ %height=0.272
%  \subfloat[\scriptsize{Depth (Z) retrieval from RGB-D data}]
%  {\label{fig:3Dprinted3Dread} \includegraphics[height=0.38\textwidth]{imageV2Localization14.png}} \\
%  \subfloat[\scriptsize{Depth (Z) estimation from nearest neighbours}]
%  {\label{fig:3Dprinted3Destimate} \includegraphics[height=0.38\textwidth]{imageV2Localization24.png}}  %\\  %0.2
%  \end{center}
%   \caption{\small{Mushrooms' center localization. In (c), the numbers in blue show the depth (Z) and the numbers in red show the distance from the sensor to the each mushroom center. We ignore the X and Y components to display for clarity except for one sample for which we showed both (X,Y,Z) position and distance (d) in meter.  The '+' sign in (b) and (c) shows the center of each mushroom.}}
%  \label{fig:3Dprinted3Dlocalization}
%\end{figure*}
%\noindent

\noindent

\noindent

\begin{figure*} [htb] %[t]%[!h]
  %\centering
  \begin{center}
   %\subfloat[\scriptsize{Original image}]
  {\includegraphics[height=0.42\textwidth]{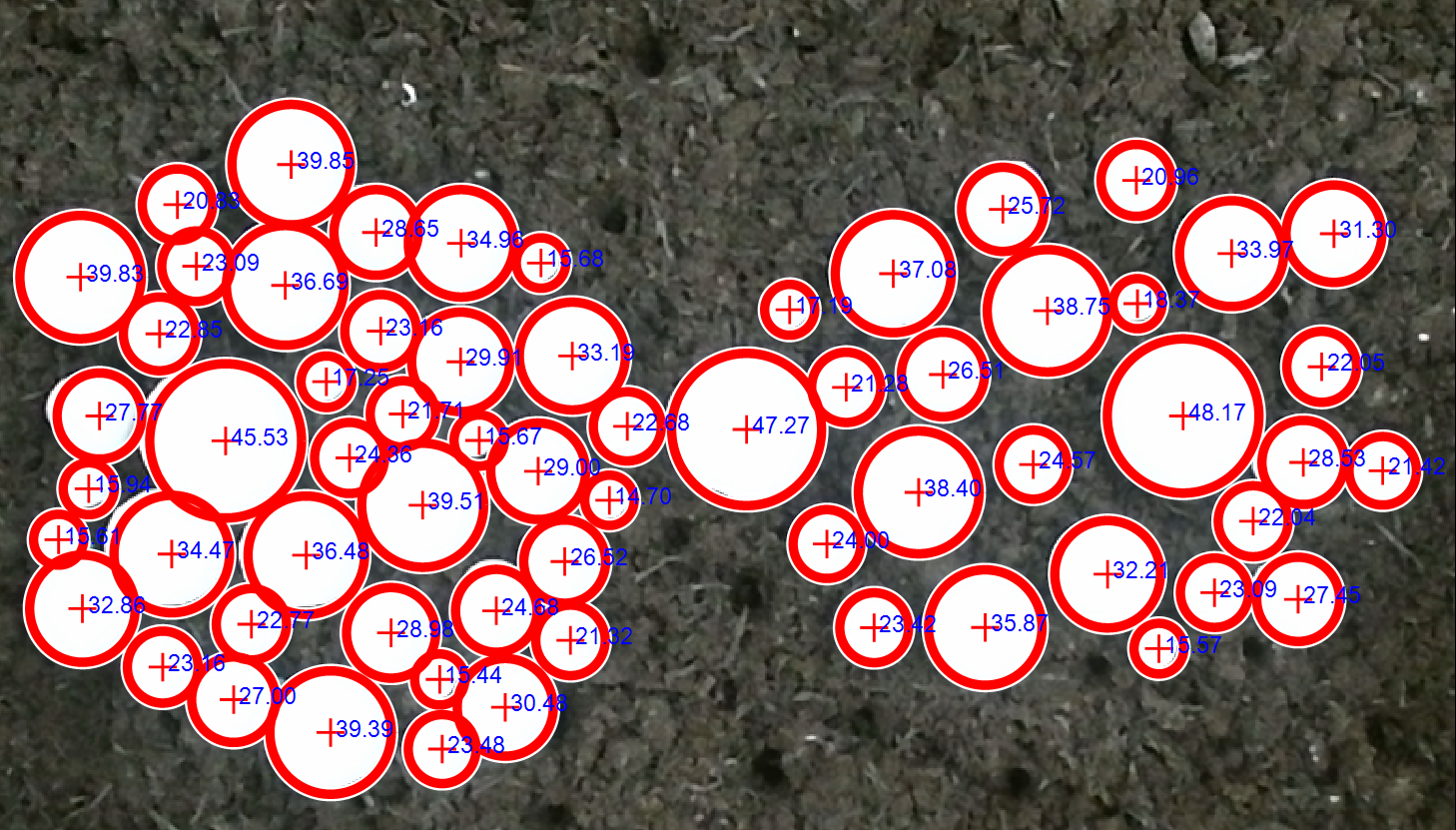}} \\ %height=0.272
  \end{center}
   \caption{\small{The estimated diameter of each mushroom in millimeter (mm) shown in blue number at each mushroom center.}}
  \label{fig:3Dprinted3Diameter}
\end{figure*}

We also estimate the diameter of each mushroom in the world coordinate system and then compute its accuracy. Using the mushrooms centers and radii computed in pixels of images, we extract the (X, Y, Z) coordinates of the point clouds obtained from the Intel RealSense. We extract the (X, Y, Z) coordinates from the opposite side of each estimated mushroom circle in image and then compute the Euclidean distance between them. This Euclidean distance is given by
\begin{equation}
Diameter = \sqrt{(X_2-X_1)^{2} + (Y_2-Y_1)^{2} + (Z_2-Z_1)^{2}}
\label{eq:Diameter}
\end{equation}
\noindent where ($X_1,Y_1,Z_1$) and ($X_2,Y_2,Z_2$) are the extracted coordinates on the opposite sides of each mushroom circle in image, basically along either row or column. We use both row and column of the circles in images and then compute the diameter in the world coordinate system by taking average of them to make robust to outliers. In case of missing depth information on the sides of the mushrooms, we estimate from the nearest available depth information. We then compare the computed diameter of each mushroom with the ground truth mushroom diameters which we measure physically in millimeters. The average offset of the estimated diameters from the true values is about -2.0mm. The computed diameters are shown in Fig.~\ref{fig:3Dprinted3Diameter} which corresponds to the images in Fig.~\ref{fig:3Dprinted3Dlocalization}. From this experiment, we observe that our algorithm shows a good performance at both detection and 3D localization of mushrooms.
% gt = [45.75, 45.5, 45.5, 34.5, 35.75, 35.25, 35.75];
% dt = [48.17, 47.27, 45.53, 37, 38.4, 35.87, 39.39];

\noindent

\begin{figure*} [htb] %[t]%[!h]
  %\centering
  \begin{center}
   \subfloat[\scriptsize{Before registration}]
  {\label{fig:3DposeBefore} \includegraphics[height=0.31\textwidth]{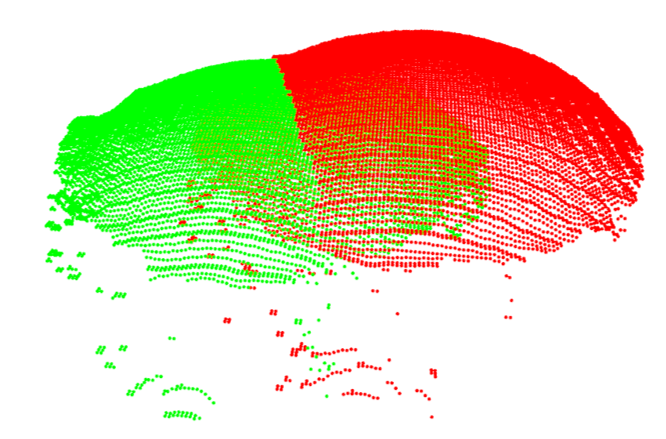}} %\\ %height=0.272
  \subfloat[\scriptsize{After registration}]
  {\label{fig:3DposeAfter} \includegraphics[height=0.31\textwidth]{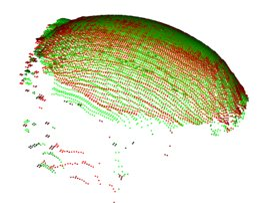}} \\%\\ %height=0.2
  \end{center}
   \caption{\small{Point clouds for model (red) and sample (green) mushrooms before and after registration.}}
  \label{fig:3Dpose1}
\end{figure*}

\begin{figure*} [htb] %[t]%[!h]
  %\centering
  \begin{center}
   \subfloat[\scriptsize{Before registration}]
  {\label{fig:3DposeBefore} \includegraphics[height=0.31\textwidth]{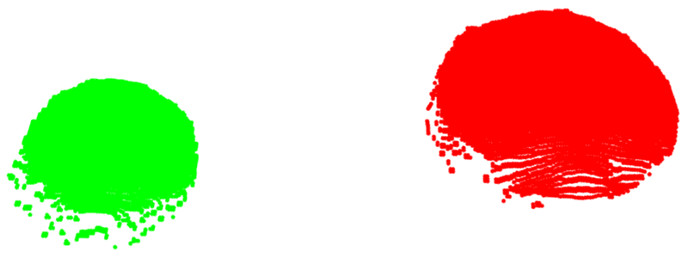}} \\ %height=0.272
  \subfloat[\scriptsize{After registration}]
  {\label{fig:3DposeAfter} \includegraphics[height=0.31\textwidth]{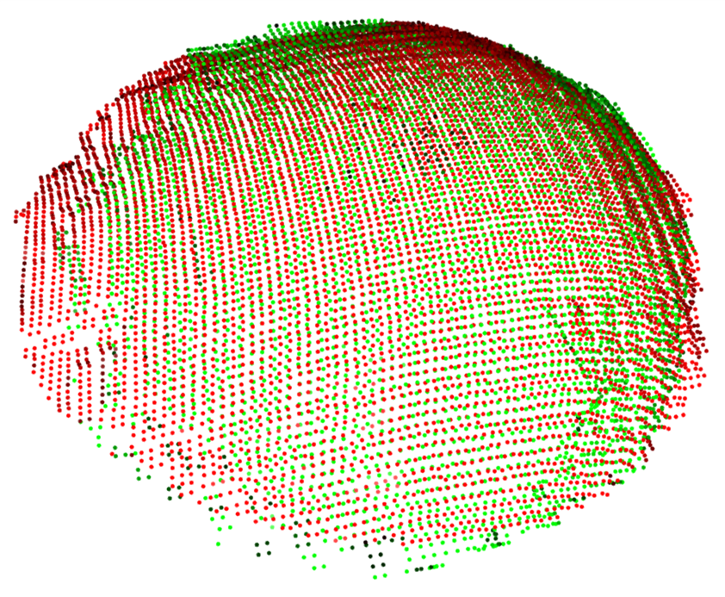}} %\\ %height=0.2
  \end{center}
   \caption{\small{Point clouds for model (red) and sample (green) mushrooms before and after registration.}}
  \label{fig:3Dpose2}
\end{figure*}

We also evaluate the performance of the proposed 3D pose estimation method. The qualitative results are shown in Fig.~\ref{fig:3Dpose1} and Fig.~\ref{fig:3Dpose2} for both before and after registration. We collected samples of mushrooms with their ground normal orientation vectors to evaluate our algorithm. We compute the angle difference ($\theta_{diff}$) between the estimated and ground truth normal orientation vectors using a dot (scalar) product as

\begin{equation}
\theta_{diff} = \cos^{-1}\bigg(\frac{\vec{v}_{gt} \boldsymbol{\cdot} \vec{v}_{est}}{\|\vec{v}_{gt}\| \|\vec{v}_{est}\|} \bigg)
\label{eq:AgnleDifference}
\end{equation}
\noindent where $\vec{v}_{gt}$ is the ground truth normal orientation vector, $\vec{v}_{est}$ is the estimated normal orientation vector obtained by multiplying the estimated rotation matrix by the unit normal ground truth vector of the mushroom model as discussed in section~\ref{sec:MushroomsPose}, $\|\vec{v}\|$ is the magnitude of a vector $\vec{v}$, $\cos^{-1}$ is an inverse cosine transform. The average angle difference between the ground truth and estimated normal orientation vectors of mushrooms is about 13.5 degrees with standard deviation of 7.5 evaluated over about 30 samples with varying angles. This is reasonable for successful mushrooms picking given the difficulty of the mushrooms orientation estimation due to their small size and textureless nature. We now have all the required parameters (X, Y, Z, Diameter, Quaternion) which will be given as input to the robot for mushrooms picking applications.

\noindent

\noindent

%\begin{figure*} [htb] %[t]%[!h]
%  %\centering
%  \begin{center}
%   \subfloat[\scriptsize{Before registration}]
%  {\label{fig:3DposeBefore} \includegraphics[height=0.31\textwidth]{pre_registration3.png}} \\ %height=0.272
%  \subfloat[\scriptsize{After registration}]
%  {\label{fig:3DposeAfter} \includegraphics[height=0.31\textwidth]{post_registration3.png}} %\\ %height=0.2
%  \end{center}
%   \caption{\small{Point clouds for model (red) and sample (green) mushrooms before and after registration.}}
%  \label{fig:3Dpose3}
%\end{figure*}
%\noindent
%From this rotation matrix R, we get a normal vector (surface normal) by multiplying it with the unit vector [0, 0, 1] as we are interested in Z-direction.

\section{Conclusion} \label{sec:Conclusion}

We have developed mushrooms detection, localization and 3D pose estimation algorithm using RGB-D data acquired from a low-cost consumer RGB-D sensor. We use RGB part (converting to grayscale for better performance) for mushrooms segmentation using active contour and detection using a circular Hough transform. Once each mushroom's center position in the RGB image is known, we then use the depth information to locate it in 3D space. In case of missing depth information at the detected center of each mushroom, we estimate from the nearest available depth information within the radius of each mushroom. We estimate the 3D pose of each mushrooms using a global registration followed by a local refine registration approach. The parameters of the mushrooms are used for robotic-picking applications. We observe that our algorithm has an interesting performance on both 3D printed mushrooms in Lab and real mushrooms from a farm in all detection, localization and orientation estimation (from the estimated 3D pose). Fusing subsequent point clouds may improve the localization and 3D pose estimation of mushrooms which can fill some holes (missing depth values) at the expense of more computational cost. Using more accurate 3D sensor such as a laser may also improve the result.

\section*{Acknowledgment}

The authors have been supported by the Innovate UK Mushroom Robo-Pic project: 103679, in conjunction with partners the University of Lincoln, Littleport Mushroom Farms LLP, Stelram Engineering Ltd and ABB Ltd.

% or


%\appendices
%% you can choose not to have a title for an appendix
%% if you want by leaving the argument blank
%\section{}

%% use section* for acknowledgement
%\section*{Acknowledgment}

\ifCLASSOPTIONcaptionsoff
  \newpage
\fi

\bibliographystyle{IEEEtran}
%\bibliography{strings,refs}
\bibliography{egbib}

\begin{thebibliography}{10}
\providecommand{\url}[1]{#1}
\csname url@samestyle\endcsname
\providecommand{\newblock}{\relax}
\providecommand{\bibinfo}[2]{#2}
\providecommand{\BIBentrySTDinterwordspacing}{\spaceskip=0pt\relax}
\providecommand{\BIBentryALTinterwordstretchfactor}{4}
\providecommand{\BIBentryALTinterwordspacing}{\spaceskip=\fontdimen2\font plus
\BIBentryALTinterwordstretchfactor\fontdimen3\font minus
  \fontdimen4\font\relax}
\providecommand{\BIBforeignlanguage}[2]{{%
\expandafter\ifx\csname l@#1\endcsname\relax
\typeout{** WARNING: IEEEtran.bst: No hyphenation pattern has been}%
\typeout{** loaded for the language `#1'. Using the pattern for}%
\typeout{** the default language instead.}%
\else
\language=\csname l@#1\endcsname
\fi
#2}}
\providecommand{\BIBdecl}{\relax}
\BIBdecl

\bibitem{GongAma15}
A.~Gongal, S.~Amatya, M.~Karkee, Q.~Zhang, and K.~Lewis, ``Sensors and systems
  for fruit detection and localization: A review,'' \emph{Computers and
  Electronics in Agriculture}, vol. 116, no. Supplement C, pp. 8 -- 19, 2015.

\bibitem{PetKoi16}
\BIBentryALTinterwordspacing
P.~Eizentals and K.~Oka, ``{3D} pose estimation of green pepper fruit for
  automated harvesting,'' \emph{Computers and Electronics in Agriculture}, vol.
  128, pp. 127 -- 140, 2016. [Online]. Available:
  \url{http://www.sciencedirect.com/science/article/pii/S0168169916307050}
\BIBentrySTDinterwordspacing

\bibitem{KapBar12}
K.~Kapach, E.~Barnea, R.~Mairon, Y.~Edan, and O.~Ben-Shahar, ``Computer vision
  for fruit harvesting robots - state of the art and challenges ahead,''
  \emph{Int. J. Comput. Vision Robot.}, vol.~3, no. 1/2, pp. 4--34, Apr. 2012.

\bibitem{SucJam16}
S.~Bargoti and J.~P. Underwood, ``Image segmentation for fruit detection and
  yield estimation in apple orchards,'' \emph{CoRR}, vol. abs/1610.08120, 2016.

\bibitem{MooCha10}
J.~Moonrinta, S.~Chaivivatrakul, M.~N. Dailey, and M.~Ekpanyapong, ``Fruit
  detection, tracking, and {3D} reconstruction for crop mapping and yield
  estimation,'' in \emph{2010 11th International Conference on Control
  Automation Robotics Vision}, Dec 2010, pp. 1181--1186.

\bibitem{CruzLuc12}
L.~Cruz, D.~Lucio, and L.~Velho, ``Kinect and {RGBD} images: Challenges and
  applications,'' in \emph{2012 25th SIBGRAPI Conference on Graphics, Patterns
  and Images Tutorials}, Aug 2012, pp. 36--49.

\bibitem{WasStr17}
O.~Wasenm{\"u}ller and D.~Stricker, \emph{Comparison of Kinect V1 and V2 Depth
  Images in Terms of Accuracy and Precision}.\hskip 1em plus 0.5em minus
  0.4em\relax Cham: Springer International Publishing, 2017, pp. 34--45.

\bibitem{YanZhaDon15}
L.~Yang, L.~Zhang, H.~Dong, A.~Alelaiwi, and A.~El~Saddik, ``Evaluating and
  improving the depth accuracy of kinect for windows v2,'' vol.~15, pp. 1--1,
  08 2015.

\bibitem{PraAld15}
J.~Prankl, A.~Aldoma, A.~Svejda, and M.~Vincze, ``{RGB-D} object modelling for
  object recognition and tracking,'' in \emph{2015 IEEE/RSJ International
  Conference on Intelligent Robots and Systems (IROS)}, Sept 2015, pp. 96--103.

\bibitem{AndJos15}
\BIBentryALTinterwordspacing
A.~Eitel, J.~T. Springenberg, L.~Spinello, M.~A. Riedmiller, and W.~Burgard,
  ``Multimodal deep learning for robust {RGB-D} object recognition,''
  \emph{CoRR}, vol. abs/1507.06821, 2015. [Online]. Available:
  \url{http://arxiv.org/abs/1507.06821}
\BIBentrySTDinterwordspacing

\bibitem{XiaYun17}
X.~Xu, Y.~Li, G.~Wu, and J.~Luo, ``Multi-modal deep feature learning for
  {RGB-D} object detection,'' \emph{Pattern Recognition}, vol.~72, no.
  Supplement C, pp. 300 -- 313, 2017.

\bibitem{SchBeh17}
M.~Schwarz, H.~Schulz, and S.~Behnke, ``{RGB-D} object recognition and pose
  estimation based on pre-trained convolutional neural network features,'' in
  \emph{2015 IEEE International Conference on Robotics and Automation (ICRA)},
  May 2015, pp. 1329--1335.

\bibitem{NguKoJeo15}
D.~D. Nguyen, J.~P. Ko, and J.~W. Jeon, ``Determination of {3D} object pose in
  point cloud with {CAD} model,'' in \emph{2015 21st Korea-Japan Joint Workshop
  on Frontiers of Computer Vision (FCV)}, Jan 2015, pp. 1--6.

\bibitem{ChoTagTuz12}
C.~Choi, Y.~Taguchi, O.~Tuzel, M.~Liu, and S.~Ramalingam, ``Voting-based pose
  estimation for robotic assembly using a {3D} sensor,'' in \emph{2012 IEEE
  International Conference on Robotics and Automation}, May 2012, pp.
  1724--1731.

\bibitem{IzadiKim11}
S.~Izadi, D.~Kim, O.~Hilliges, D.~Molyneaux, R.~Newcombe, P.~Kohli, J.~Shotton,
  S.~Hodges, D.~Freeman, A.~Davison, and A.~Fitzgibbon, ``Kinectfusion:
  Real-time {3D} reconstruction and interaction using a moving depth camera,''
  in \emph{Proceedings of the 24th Annual ACM Symposium on User Interface
  Software and Technology}, ser. UIST '11.\hskip 1em plus 0.5em minus
  0.4em\relax New York, NY, USA: ACM, 2011, pp. 559--568.

\bibitem{EndCre14}
F.~Endres, J.~Hess, J.~Sturm, D.~Cremers, and W.~Burgard, ``{3D} mapping with
  an {RGB-D} camera,'' \emph{IEEE Transactions on Robotics}, vol.~30, no.~1,
  pp. 177--187, Feb 2014.

\bibitem{Mas13}
A.~Masoudian, ``Computer vision algorithms for an automated harvester,''
  Electronic Thesis and Dissertation Repository. 1804, The University of
  Western Ontario, 2013.

\bibitem{ChanVese01}
T.~F. Chan and L.~A. Vese, ``Active contours without edges,'' \emph{IEEE
  Transactions on Image Processing}, vol.~10, no.~2, pp. 266--277, Feb 2001.

\bibitem{MarBauAlv14}
P.~Márquez-Neila, L.~Baumela, and L.~Alvarez, ``A morphological approach to
  curvature-based evolution of curves and surfaces,'' \emph{IEEE Transactions
  on Pattern Analysis and Machine Intelligence}, vol.~36, no.~1, pp. 2--17, Jan
  2014.

\bibitem{AthKer99}
T.~Atherton and D.~Kerbyson, ``Size invariant circle detection,'' \emph{Image
  and Vision Computing}, vol.~17, no.~11, pp. 795 -- 803, 1999.

\bibitem{ZhoParKol18}
Q.-Y. Zhou, J.~Park, and V.~Koltun, ``{Open3D}: {A} modern library for {3D}
  data processing,'' \emph{arXiv:1801.09847}, 2018.

\bibitem{RusBloBee09}
R.~B. Rusu, N.~Blodow, and M.~Beetz, ``Fast point feature histograms (fpfh) for
  {3D} registration,'' in \emph{2009 IEEE International Conference on Robotics
  and Automation}, May 2009, pp. 3212--3217.

\bibitem{HolIchTom15}
D.~Holz, A.-E. Ichim, F.~Tombari, R.~B. Rusu, and S.~Behnke, ``Registration
  with the point cloud library a modular framework for aligning in {3-D},''
  \emph{Ieee Robotics and Automation Magazine}, vol.~22, no.~4, pp. 15.
  110--124, 2015.

\bibitem{ZhoParKol16}
Q.-Y. Zhou, J.~Park, and V.~Koltun, ``Fast global registration,'' in
  \emph{Computer Vision -- ECCV 2016}, B.~Leibe, J.~Matas, N.~Sebe, and
  M.~Welling, Eds.\hskip 1em plus 0.5em minus 0.4em\relax Cham: Springer
  International Publishing, 2016, pp. 766--782.

\bibitem{pomColSie13}
\BIBentryALTinterwordspacing
F.~Pomerleau, F.~Colas, R.~Siegwart, and S.~Magnenat, ``{Comparing ICP variants
  on real-world data sets Open-source library and experimental protocol},''
  \emph{{Autonomous Robots}}, vol.~34, no.~3, pp. 133--148, 2013. [Online].
  Available: \url{https://hal.archives-ouvertes.fr/hal-01143458}
\BIBentrySTDinterwordspacing

\bibitem{RusLev01}
S.~Rusinkiewicz and M.~Levoy, ``Efficient variants of the {ICP} algorithm,'' in
  \emph{Proceedings Third International Conference on {3-D} Digital Imaging and
  Modeling}, May 2001, pp. 145--152.

\bibitem{BesMcK92}
P.~J. Besl and N.~D. McKay, ``A method for registration of {3-D} shapes,''
  \emph{IEEE Transactions on Pattern Analysis and Machine Intelligence},
  vol.~14, no.~2, pp. 239--256, Feb 1992.

\bibitem{CheMed92}
\BIBentryALTinterwordspacing
Y.~Chen and G.~Medioni, ``Object modelling by registration of multiple range
  images,'' \emph{Image and Vision Computing}, vol.~10, no.~3, pp. 145 -- 155,
  1992, range Image Understanding. [Online]. Available:
  \url{http://www.sciencedirect.com/science/article/pii/026288569290066C}
\BIBentrySTDinterwordspacing

\bibitem{ParZhoKol17}
J.~Park, Q.~Zhou, and V.~Koltun, ``Colored point cloud registration
  revisited,'' in \emph{2017 IEEE International Conference on Computer Vision
  (ICCV)}, Oct 2017, pp. 143--152.

\end{thebibliography}


%\vfill


\end{document}